**3D Object Detection and High-Resolution Traffic Parameters Extraction Using Low-Resolution LiDAR Data**

**Linlin Zhang**
Ph.D. Student
Department of Civil and Environmental Engineering
University of Missouri-Columbia, Columbia, MO, USA, 65201
Email: lz5f2@mail.missouri.edu

**Xiang Yu**
Ph.D. Student
Department of Civil and Environmental Engineering
University of Missouri-Columbia, Columbia, MO, USA, 65201
CFS Engineers, Kansas City, MO, USA, 64131
Email: xytm4@mail.missouri.edu

**Armstrong Aboah**
Assistant Research Professor
Department of Civil and Architectural Engineering and Mechanics
University of Arizona, Tucson, AZ, USA, 85721
Email: aaboah@arizona.edu

**Yaw Adu-Gyamfi**
Associate Professor
Department of Civil and Environmental Engineering
University of Missouri-Columbia, Columbia, MO, USA, 65201
Email: adugyamfiy@missouri.edu

Word Count: 6,680 words + 3 tables (250 words per table) = 7,430 words

Submitted for consideration for presentation at the 103rd Annual Meeting of the Transportation Research Board, January 2024

Submitted Date: August 1, 2023

**ABSTRACT**

Traffic volume data collection is a crucial aspect of transportation engineering and urban planning, as it provides vital insights into traffic patterns, congestion, and infrastructure efficiency. Traditional manual methods of traffic data collection are both time-consuming and costly. However, the emergence of modern technologies, particularly Light Detection and Ranging (LiDAR), has revolutionized the process by enabling efficient and accurate data collection. Despite the benefits of using LiDAR for traffic data collection, previous studies have identified two major limitations that have impeded its widespread adoption. These are the need for multiple LiDAR systems to obtain complete point cloud information of objects of interest, as well as the labor-intensive process of annotating 3D bounding boxes for object detection tasks. In response to these challenges, the current study proposes an innovative framework that alleviates the need for multiple LiDAR systems and simplifies the laborious 3D annotation process. To achieve this goal, the study employed a single LiDAR system, that aims at reducing the data acquisition cost and addressed its accompanying limitation of missing point cloud information by developing a Point Cloud Completion (PCC) framework to fill in missing point cloud information using point density. Furthermore, we also used zero-shot learning techniques to detect vehicles and pedestrians, as well as proposed a unique framework for extracting low to high features from the object of interest, such as height, acceleration, and speed. Using the 2D bounding box detection and extracted height information, this study is able to generate 3D bounding boxes automatically without human intervention.

## INTRODUCTION

The collection of traffic volume data is an essential component of transportation engineering and urban planning. It provides valuable insights into the flow of vehicular traffic and contributes to the development of efficient transportation systems. In the past, traffic data collection methods were predominantly manual and beset with certain limitations. Roadside manual counts, for instance, were labor-intensive and susceptible to human error. Similarly, pneumatic tubes and inductive loop detectors, although used, lacked comprehensive traffic flow information. Additionally, reliance on manual surveys introduced biases and inaccuracies in the collected data. The advent of modern, non-invasive sensing technologies has, however, revolutionized traffic data collection approaches, mitigating many of these limitations. Notably, LiDAR (Light Detection and Ranging) technology has emerged as a transformative solution, offering higher accuracy, real-time monitoring, and extensive coverage (*1*). The use of this technology facilitates comprehensive traffic data collection, including vehicle types, speed, occupancy, and 3D mapping of road networks. Moreover, its non-intrusive nature and enhanced safety measures for pedestrians and cyclists render LiDAR an appealing choice for traffic volume data collection. Nevertheless, it is imperative to acknowledge that the cost of a single LiDAR unit is substantial, presenting a financial challenge for researchers and industry practitioners seeking to harness this technology for data collection.

Prior studies employing LiDAR for traffic data collection and analysis have identified two significant constraints that have hindered its extensive implementation and adoption. Firstly, obtaining a complete point cloud representation of an object of interest often necessitates the use of two LiDAR systems, significantly raising the overall data acquisition cost. While acknowledging the paramount importance of high-quality data collection, the economic feasibility of the acquisition process emerges as a critical concern. This raises the question of whether comparable or equivalent outcomes can be achieved by utilizing a single LiDAR system. Secondly, employing LiDAR data for vehicle detection and classification models necessitates labor-intensive and time-consuming 3D bounding box annotations, posing challenges in terms of resource allocation and analysis efforts. The process of annotating 3D bounding boxes proves to be a significant challenge, adding complexity and increasing the effort required for subsequent analysis. Addressing these challenges constitutes the primary objective of the current study.

To overcome these limitations, the present study aims to propose an innovative solution that reduces the dependence on multiple LiDAR systems for obtaining full point representations of objects of interest while maintaining data quality. Additionally, the study explores the application of zero-shot learning techniques to streamline the arduous 3D bounding box annotation process necessary for developing object detection and classification models using LiDAR data. To achieve these objectives, a single LiDAR technology was installed at an intersection in Columbia-Missouri, enabling data collection for both vehicular traffic and pedestrians. By employing a stationary LiDAR setup, moving objects (vehicles and pedestrians) were identified by estimating the background of the bird's eye view (BEV) images and subtracting it from each BEV image to observe the moving objects. To address issues of missing point cloud information due to partial occlusion as a result of using a single LiDAR system, the current study developed a framework called Point Cloud Completion (PCC) framework to fill in any missing point cloud information based on point density. Subsequently, the study leverages a zero-shot learning technique by employing the Segment Anything Model (SAM) (*2*) to detect and classify moving objects specifically vehicles and pedestrians. To facilitate traffic counting at the intersection, the detected bounding boxes are tracked using the Bytetrack algorithm (*3*). Furthermore, the study devises a framework to transform the 2D bounding boxes, estimated by the SAM model, into 3D bounding boxes, incorporating object height information derived from the LiDAR data. By employing these advanced methodologies, the study aims to achieve more accurate and efficient traffic data collection and analysis, advancing the field of transportation engineering and urban planning.

The contributions of this research lie in its potential to improve data acquisition affordability and streamline data annotation, ultimately fostering advancements in transportation engineering and urban planning practices. The study makes significant contributions in three key areas as follows.

1. We propose an innovative and effective framework for obtaining full point cloud representations of partially occluded objects using just a single LiDAR system, thus addressing the limitation of requiring multiple LiDAR units for comprehensive data acquisition.

2. We introduce a novel framework for generating 3D bounding box annotations without relying on human annotators. This development is crucial in streamlining the labor-intensive and time-consuming process of annotating LiDAR data for traffic detection and classification models.

3. The study pioneers the application of zero-shot learning techniques for vehicle and pedestrian detection, classification, and counting.

The rest of the paper is organized as follows. Section two presents a discussion of relevant literature. The data and data collection process are presented in section three. Section four discusses the methodology for the study. The results and discussion of the study are presented in section five. Section six discusses the conclusions, limitations, and recommendations of this study.

**RELATED WORKS**

In this section, the primary emphasis of the reviewed papers centers around two principal domains: object detection and classification of LiDAR Bird-Eye View (BEV) images and Zero-Shot Learning (ZSL) techniques.

**LiDAR Bird-Eye View (BEV) Object Detection and Classification**

The convolutional neural network (CNN) architecture has been widely used in feature extraction and aggregation in BEV object detection (*4*). Beltran et al. employed VGG-16 architecture (*5*) for feature extraction and utilized Faster R-CNN (*6*) for object detection and orientation estimation on the BEV images. However, an additional step is to estimate the object height from the ground plan to construct 3D bounding boxes (*7*). Barrera et al. conducted further research on (*7*) and proposed a two-stage oriented-boxes estimation approach to eliminate the need for the extra step. The first stage utilized the Regional Proposal Network (RPN) to generate axis-aligned proposals. While the second stage used two fully connected layers for tasks including classification, discretized yaw classification, and axis-aligned box regression. Instead of VGG-16 architecture (*5*), ResNet-50 (*8*) was adopted for feature extraction because of the computation speed and accuracy (*9*). Lee et al. proposed a novel approach to discriminate and track static objects and dynamic objects by utilizing the paired two consecutive BEV images as input. Then, an optical flow network is adopted to obtain a 2-D BEV flow grid (*10*).

To enhance the detection results, the fusion of BEV with other views from LiDAR or cameras was investigated. Zhou et al. presented a multi-view fusion framework that combined the LiDAR BEV and LiDAR perspective views (*11*). Wang et al. employed a novel layer, named sparse non-homogeneous pooling layer, to fuse the features from LiDAR BEV and camera front view. Subsequently, this layer is integrated into the CNN architecture (*12*). To address the challenges of object detection in adverse weather conditions, Yang et al. focused on the fusion of features from radar range-azimuth heatmap and from LiDAR point cloud. Their approach was capable of accurately performing 2D bounding box detection in BEV images (*4*).

**Zero-Shot Learning (ZSL)**

Zero-shot learning (ZSL) is an advanced machine learning technique that enables a model to classify objects from classes that were not observed during its training phase, without requiring specific training data for those specific classes. This is accomplished through the use of a foundation model. A foundation model (*13*) is a model that has been pre-trained using a large dataset containing labeled information to gain comprehensive visual and semantic representations of objects, enabling it to classify novel objects originating from previously unseen classes. ZSL, like any other machine learning technique, is beset with some limitations which include substantial bias between the seen and unseen class in the training dataset (*14*), the presence of domain shift (*15*), and the hubness problem (*16*). Nevertheless, ZSL has also achieved great success in recent years. In recent years, foundation models (*13*) for ZSL have achieved significant success in linguistic (*17, 18*) and visual (*19–21*) tasks respectively. More recently the eruption of a new visual foundation model named the "Segment Anything Model" (SAM), trained on an extensive dataset known as SA1B, has brought about a paradigm shift in the field (*2*). The introduction of this model has revolutionized the field of ZSL, and subsequent research efforts have utilized SAM's exceptional zero-shot capability to address various 2D vision tasks, such as medical image processing (*22–24*) and camouflaged object segmentation (*25–27*). Remarkably, in the domain of civil engineering, Mohsen et al. have effectively leveraged SAM to facilitate civil infrastructure defect assessment (*28*), while Zhou et al. have utilized the model to segment any moving object in conjunction with a moving ego vehicle (*29*). Moreover, Di Wang et al. have demonstrated SAM's suitability for constructing an extensive Remote Sensing segmentation dataset by acquiring pixel-level annotations for remote sensing images (*30*). The versatility of SAM has further been exemplified in the domain of autonomous driving through the adept application by Xinru et al. (*31*). Although SAM presents great power on some 2D vision tasks, its capabilities on 3D vision tasks have not been explored widely. As such this study seeks to explore using the SAM model for 3D vision tasks.

Building on these successful studies, we applied a well-known Zero-shot Learning model (SAM) and other computer vision techniques on low-resolution LiDAR Bird-Eye View (BEV) images to achieve the construction of 3D bounding boxes for detected objects and extract various traffic-related high-resolution parameters.

**DATA COLLECTION**

For this work, the 64-channel Ouster LiDAR sensor (operation mode 2048 x 10 hz) was mounted on a vehicle positioned at the northwest corner of the intersection of E Broadway & N 9th Street, Columbia, Missouri. This intersection is signalized with a four-leg configuration and marked crosswalks. On E Broadway, the posted speed limit is 20 mph and angled parking stalls are installed along the street. For N 9th Street, the posted speed limit is 25 mph, and parallel parking spaces are installed along the street. **Figure 1** shows how the LiDAR sensor was installed and the data collection location.

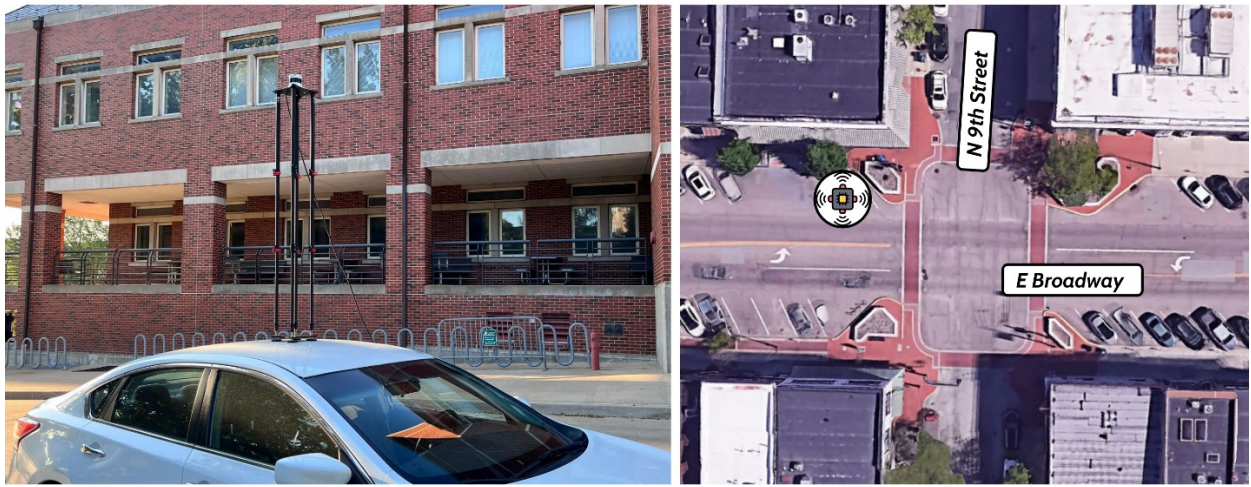

**Figure 1.** The LiDAR Sensor Installation and Data Collection Location

**METHODOLOGY**

In this paper, we proposed a novel methodological framework shown in **Figure 2** for 3D object detection, tracking, and high-resolution parameter extraction from LiDAR data. The framework encompasses five main sequential steps. Initially, we transform the 3D point clouds to 2D point clouds (BEV plane) for easier analysis. Subsequently, we developed a unique object detection pipeline for detecting vehicles and pedestrians from the BEV images. This pipeline consists of 1) the determination of static and moving objects from the BEV images using the background subtraction technique, 2) completing partially occluded objects with missing point cloud information using the PCC algorithm before subjecting them to the SAM, and 3) employing the SAM model for Zero-Shot Learning (ZSL) object detection of the identified moving objects. To ensure the detection of accurate and distinct bounding boxes for moving objects obtained in the previous step, a non-maximum suppression (NMS) algorithm is implemented to remove overlapping bounding boxes, and the Bytetrack algorithm (*32*) is utilized to eliminate redundant bounding boxes. In the subsequent step, an approach is devised to generate the 3D representation of the detected 2D moving objects, incorporating height information from the LiDAR system. Finally, high-resolution traffic flow parameters, such as counts, speed, and acceleration, are accurately extracted using data acquired from a single stationary-mounted LiDAR sensor. Subsequent sections provide a comprehensive and elaborate description of each step of the framework.

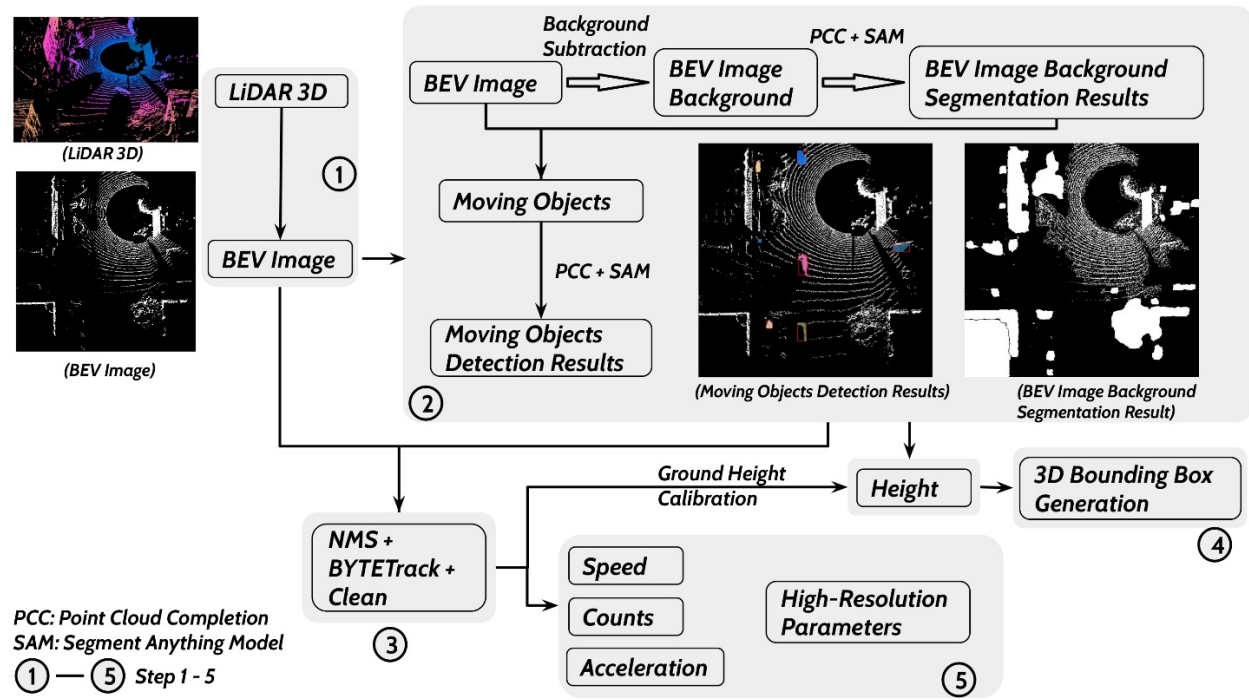

**Figure 2**. Proposed Pipeline

**3D-2D Point Cloud Data Projection**

To facilitate analysis of the point cloud dataset, the study transformed the 3D point clouds into 2D point clouds (BEV plane). This procedural step was crucial to overcome the challenges associated with 3D object detection tasks as well as obtain a macro transportation layout map within the designated Region of Interest (ROI). The transformation process involved collapsing the vertical plane (z plane) while retaining the BEV plane (x and y planes) enabling a comprehensive representation of the traffic situation within the ROI. **Figure 3** illustrates the 3D point cloud and its corresponding 2D BEV image.

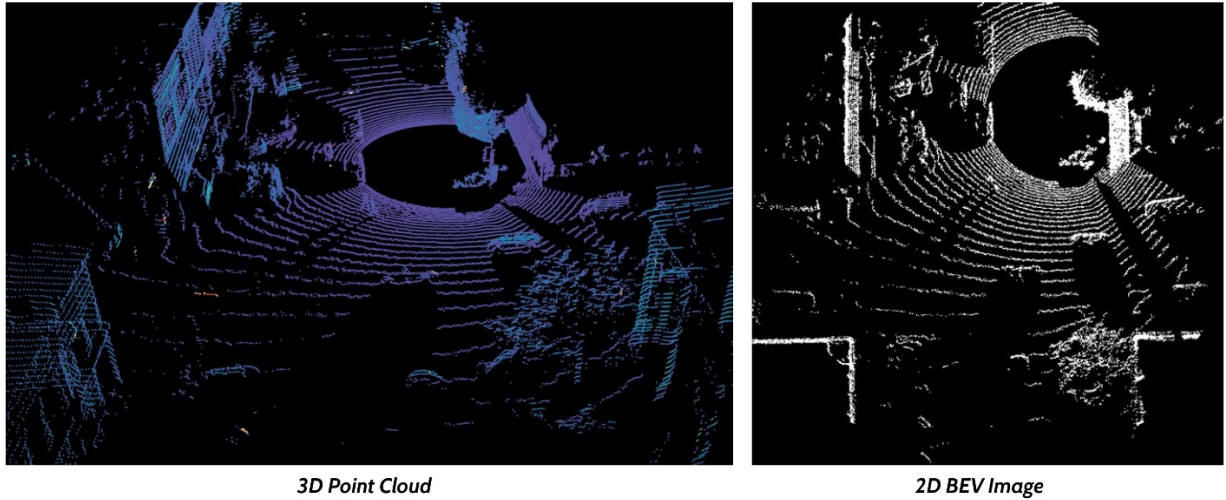

**3D Point Cloud**  **2D BEV Image**

**Figure 3**. 3D Point Cloud and 2D BEV Image at the Same Intersection

**Moving Object Detection via Zero-Shot Learning**

The second step of our proposed methodology involves the detection of moving objects, specifically vehicles and pedestrians. To accomplish this step, two major processes were utilized. That is background subtraction and zero-shot object detection technique.

*Background Subtraction*

Background subtraction is a frequently utilized algorithm in the field of computer vision and image processing, designed to isolate moving objects from a static background. This is achieved by extracting the stationary background information from a sequence of images or videos, enabling the algorithm to detect changes caused by moving objects effectively. By employing this technique, efficient analysis and understanding of dynamic environments become possible. In this study, we applied the background subtraction technique to a series of BEV images to isolate the foreground from the background. Though effective, we were faced with the challenges of missing point cloud information due to partial or full occlusion of objects of interest. We subsequently addressed this problem using the proposed point cloud completion (PCC) algorithm (see **Figure 5**) to complete these missing point clouds based on point density. The PCC algorithm also aims to enhance the completeness of the background region. The resultant image shown in **Figure 4** (d) is passed onto the object detection step.

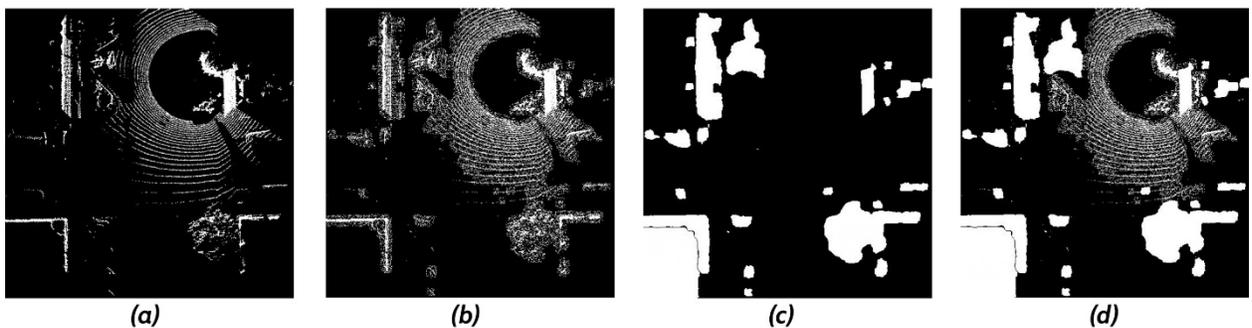

*(a)*  *(b)*  *(c)*  *(d)*

**Figure 4**. Post-Processing of Background Subtraction with PCC and SAM. (a) the background subtraction result using a series of BEV images, (b) PCC on raw background subtraction, (c) SAM-based segmentation on (b), (d) PCC on SAM-based segmentation (combination of b and c)

| **Algorithm** Point Cloud Completion |
|---|
| 1: **Input:** Image $X$ of size $N \times N$, filter size $n$, threshold of density ratio $\boldsymbol{\rho}$, noise rate $\boldsymbol{\alpha}$ |
| 2: **Function** PCC($\boldsymbol{X, n, \rho, \alpha}$) |
| 3:    let $\boldsymbol{d_s} \leftarrow$ density of area S |
| 4:    let $\boldsymbol{X_{noise}} \leftarrow$ copy of $\boldsymbol{X}$ |
| 5:    **for** $i = n$ to $M - n$ **do** |
| 6:      **for** j= $n$ to N $- n$ **do** |
| 7:        $\boldsymbol{img\_win} \leftarrow X[\boldsymbol{i - n: i + n, \ j - n: j + n}]$ |
| 8:        **if** $\boldsymbol{d_{img\_win}} > 2n \times 2n \times \rho$ **then** |
| 9:          $\boldsymbol{X_{noise}}[\boldsymbol{i - n: i + n, \ j - n: j + n}] \leftarrow$ **salt-and-pepper noise on a rate of** $\boldsymbol{\alpha}$ |
| 10:   return $\boldsymbol{X_{noise}}$ |

**Figure 5.** Algorithm of Point Cloud Completion

*Zero-Shot Learning and Object Detection*

   In our object detection step, we employ Zero-Shot Learning (ZSL), a machine learning technique that enables the model to classify objects from classes it hasn't encountered during its training phase, without the need for specific training data for those particular classes. Within this context, we leverage the ZSL technique to detect moving objects from the Bird's Eye View (BEV) images. We utilized the SAM model for our ZSL object detection. Furthermore, the PCC algorithm applied in the preceding step played a vital role in addressing the partial occlusion issue arising from the single LIDAR setup. It successfully fills small, occluded gaps of an object of interest with salt-and-pepper noise, effectively bridging the missing parts and forming a complete object area. This enhancement substantially aids the SAM model in accurately detecting the moving objects in the BEV images, thereby improving the overall object detection process. **Figure 6** illustrates the moving object detection result utilizing PCC and SAM model. For visual clarity, the color of the moving objects is randomly assigned in the figure.

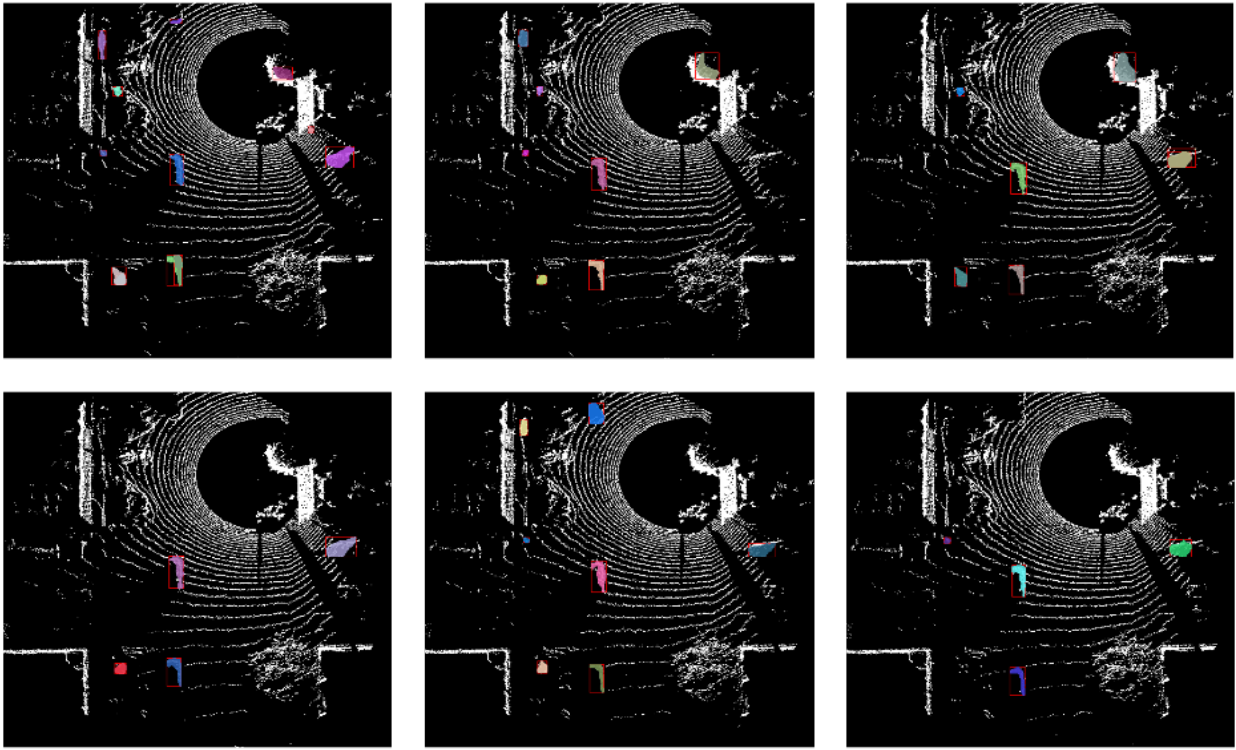

**Figure 6.** Moving Object Detection Using PCC and Zero-Shot Learning Techniques

## Moving Object Trajectory Construction

*Moving Object Tracking*

In this section, two computer vision techniques, Non-Maximum Suppression (NMS) and BYTEtrack (*3*), are adopted to accomplish the tracking task. The NMS is widely used to retain the most relevant bounding box from a group of overlapping bounding boxes. The BYTEtrack (*3*) is a well-known start-of-the-art multi-object tracking algorithm that can efficiently handle occlusion situations. The reason for adopting BYTEtrack is its advantage of flexible matching matrix selection, allowing the option to switch between simple Intersection over Union (IOU) or feature similarity for matching purposes. Due to the binary nature of BEV images, the feature extractor matching is disregarded, and only IOU matching is employed in this section. The exclusive reliance on detected bounding boxes enhances the efficiency and speed during tracking operation, making it more effective.

*Post-Processing of Tracks*

After object tracking, the subsequent step is to filter out the fragmented or unreasonable trajectories which are normally caused by occlusion and missed object detection. The filtering criteria are as follows: 1) checking frequency composition since the trajectories are formed by the frequency of the same object's appearance. The trajectories with insufficient or excessive form frequencies are removed; 2) comparing the trajectory's length with the spatial distance between its start and end points. If there is a substantial difference between the two. The trajectory is removed; 3) checking the bounding box with any disproportionate aspect ratios, for example, if a bounding box exhibits a width that is significantly larger than the height while covering a relatively small area, it is considered to be removed. Through applying these filtering criteria, some error trajectories are removed which will enhance the accuracy of high-resolution parameters estimation. Finally, four high-resolution parameters- counts, speed, acceleration, and height, can be extracted using the cleaned tracks. Subsequently, the focus shifts towards the extraction of height information from the 3D view to create 3D bounding boxes, and count, speed, and acceleration from 2D BEV representation.

**3D Bounding Box Estimation**

Even though LiDAR can provide extremely accurate dimensions information about objects. In our case, the height of all the points in the point cloud is measured as relative height to the LiDAR sensor coordinate system. For the same vehicle in motion, the relative height varies due to the slopes of the road, or the LiDAR sensor pose. Thus, to obtain the vehicle's ground truth height, the calibration of ground height is necessary. Since the LiDAR sensor is mounted on a vehicle in a stationary position, not while moving with the vehicle, accurately obtaining ground height is achievable.

**Figure 7** shows the perspective view of the LiDAR point cloud. It can be noticed that the relative height of vehicles varies with position. These variations are attributed to factors such as the slopes of a road or the LiDAR sensor pose during data collection.

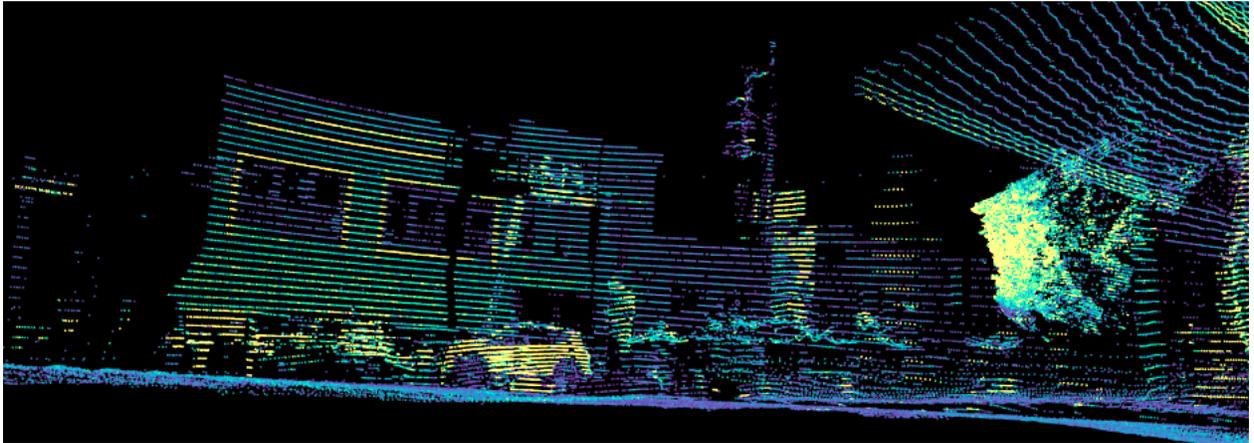

**Figure 7.** The Perspective View of LiDAR Point Cloud

In Step 2, the stationary background of the BEV image is first extracted, followed by the application of the PCC technique and SAM. Using this image, the coordinates (x, y) of the surrounding road surface and buildings can be determined based on the coordinates of the white pixels (representing segmented areas and point cloud area) in the image. Subsequently, the height (z) information of the surrounding road surface and buildings can be extracted from the 3D point cloud with coordinates of x and y.

However, the LiDAR point cloud did not cover all the areas of the surrounding roadway surface. Fortunately, in the Region of Interest (ROI), the road surface slopes exhibit minimal variations. To address this limitation and obtain the ground height, an interpolation technique is utilized to estimate height values in areas where the points are not present in the 2D background BEV image. In **Figure 8**, the right image represents the stationary background BEV image with the application of the PCC technique and SAM; the left image represents the ground height map with the interpolation technique applied.

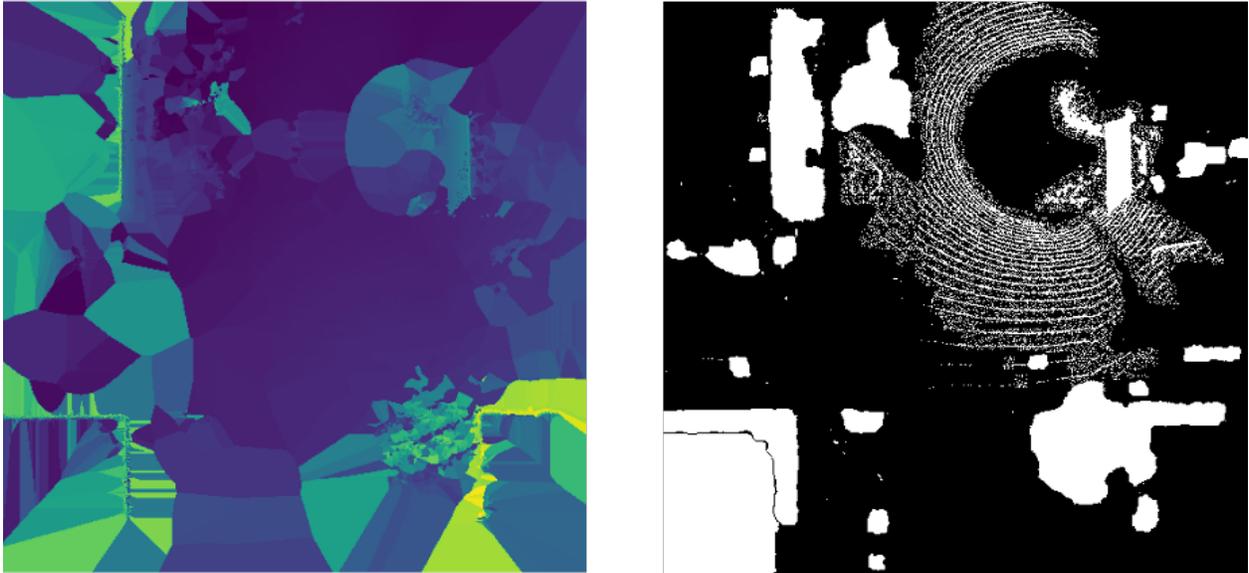

**Figure 8.** Interpolated Ground Height Map using 2D background BEV Image with PCC and Zero-Shot Learning SAM and the 2D BEV Background

The heights of objects are determined by the difference between the heights of the objects relative to the LiDAR sensor and the height of the ground. To eliminate some potential height outliers, the 95th percentile of height information within the specific bounding box of each object is utilized and regarded as the object's real height. The final height of the object bounding box is calculated by adding a small offset to the object's real height. **Figure 9** illustrates the moving object with its predicted 3D bounding box.

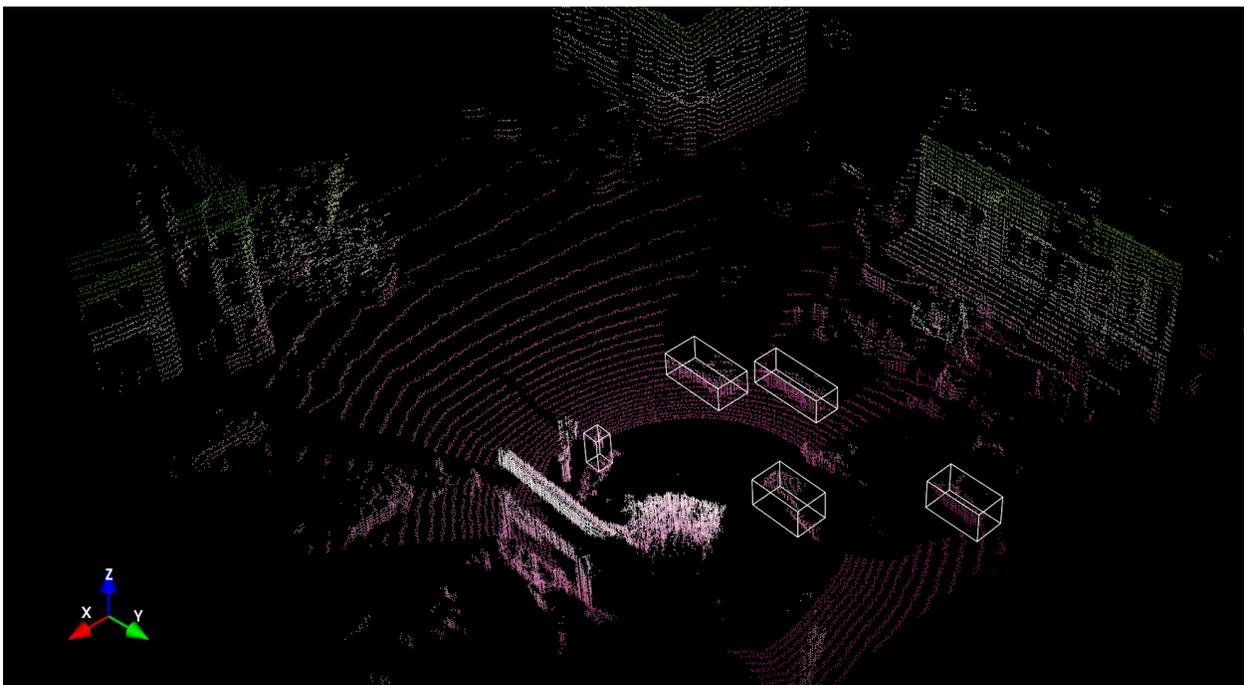

**Figure 9.** Moving Objects With its Predicted 3D Bounding Boxes

**High-Resolution Parameters Calculation**

The 3D bounding box information is integrated with the tracking result of each object, and then the Gaussian Filter is employed to enhance trajectory smoothness, as shown in **Figure 10**. By utilizing the smoothed trajectory in meters with a frame rate of 10 fps, high-resolution parameters such as speed and acceleration can be directly calculated. Additionally, the BEV image provides a clear distinction between pedestrians and vehicles, allowing classification based on the area of the bounding box in the BEV representation.

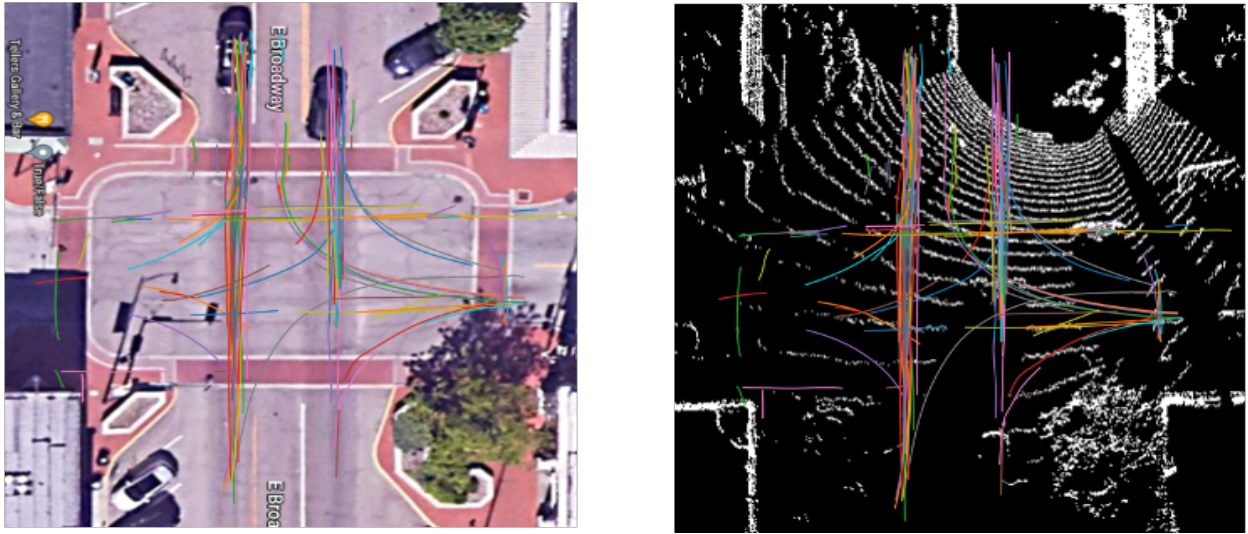

**Figure 10.** Trajectories of Detected Objects After Gaussian Filter

**RESULT AND DISCUSSION**

The accuracy of the predicted 3D bounding boxes is evaluated by comparing them with the corresponding ground truth of the 3D bounding boxes. The high-resolution parameters are evaluated by comparing statistical data with the real traffic conditions at the intersection.

**Evaluation of 3D Bounding Box**

*3D Bounding Box Ground Truth*

Due to the rare availability of open-source LiDAR point cloud data with ground truth annotations at intersections, the tool named Ground Truth Labeler in MATLAB is utilized to annotate the 3D bounding boxes of detected objects. This visualization tool offers various perspectives (see **Figure 11**) on the LiDAR point cloud, such as BEV, top-view, side-view, and front-view, which aids in precise annotation and labeling. The output of each annotated result is represented as a matrix like $[x_{ctr}, y_{ctr}, z_{ctr}, x_{len}, y_{len}, z_{len}, x_{rot}, y_{rot}, z_{rot}]$, where $[x_{ctr}, y_{ctr}, z_{ctr}]$ means the center of the 3D bounding box in 3D coordinates; $[x_{len}, y_{len}, z_{len}]$ denotes the dimensions of the 3D bounding box; $[x_{rot}, y_{rot}, z_{rot}]$ represents the rotation of the 3D bounding box. Therefore, in our study, the annotated 3D bounding box information directly from LiDAR point cloud in MATLAB is regarded as ground truth.

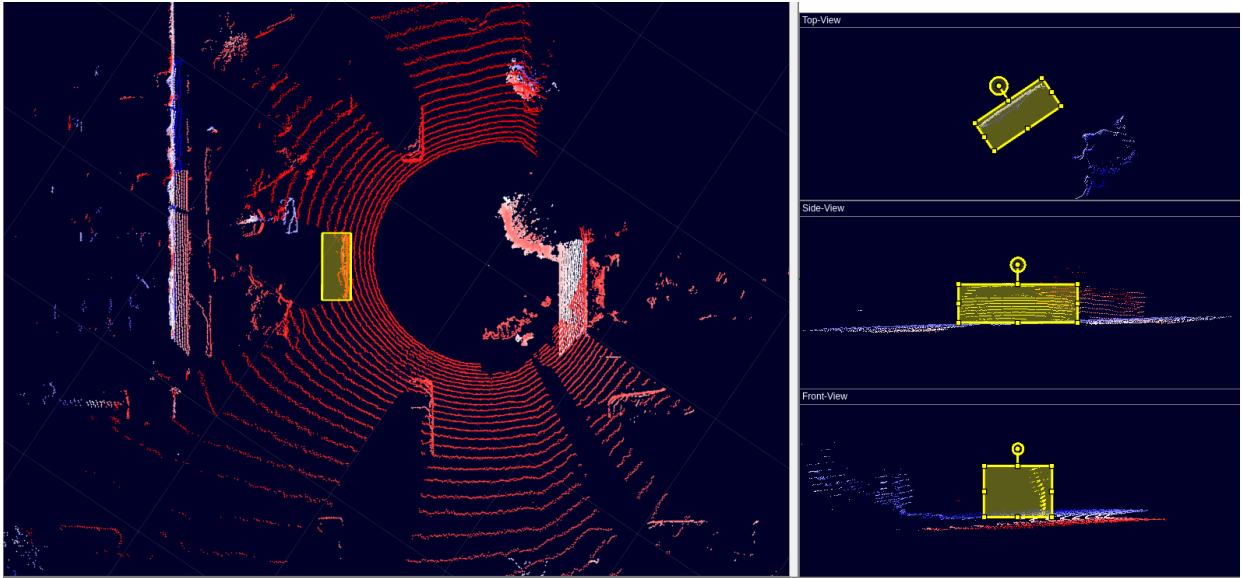

**Figure 11.** The Ground Truth Extraction with the Tool Ground Truth Labeler in MATLAB

We assess the 3D bounding boxes generated using low-resolution LiDAR data from two perspectives. Firstly, we employ the 3D Intersection over Union (IoU) algorithm, which calculates the ratio of the intersection volume over the union volume. Additionally, we analyze the absolute differences in center coordinates, as well as the length, height, and width of the predicted and ground truth bounding boxes to further evaluate the performance. Due to the low-resolution of the LiDAR sensor used and the fact that the point cloud becomes sparse with increasing distance, this brings some challenges to the generation of 3D bounding boxes. We have narrowed down the evaluation distance range to ensure the accuracy of the ground truth values. Therefore, we perform comparative analysis for both vehicles and pedestrians within two distinct distance ranges: within a radius of 15 meters and beyond 15 meters.

**Table 1.** Accuracy Evaluation Results for the Detected 3D Bounding Boxes Against the Ground Truth

| Class | Range | Volume IoU | Front IoU | BEV IoU | Side IoU |
|---|---|---|---|---|---|
| Overall | Overall | 0.34 | 0.45 | 0.51 | 0.46 |
| | <=15 m | 0.40 | 0.50 | 0.55 | 0.57 |
| | >15 m | 0.30 | 0.45 | 0.47 | 0.43 |
| Vehicle | Overall | 0.41 | 0.53 | 0.60 | 0.55 |
| | <=15 m | 0.43 | 0.52 | 0.60 | 0.58 |
| | >15 m | 0.37 | 0.55 | 0.59 | 0.47 |
| Pedestrian | Overall | 0.23 | 0.43 | 0.27 | 0.43 |
| | <=15 m | 0.32 | 0.47 | 0.43 | 0.53 |
| | >15 m | 0.15 | 0.35 | 0.16 | 0.39 |

**Table 1** illustrates the accuracy evaluation results for the predicted 3D bounding boxes against the ground truth of the vehicle. According to **Table 1**, the overall average IoU value between the predicted bounding boxes and the ground truth is 0.34. To investigate the specific factors contributing to the low IoU scores, we further calculated the 2D IoU values for three different views- BEV, front view, and side view.

For the class Vehicle, the IoU score for BEV exhibits a notably higher value compared to the front view and side view. This indicates a more accurate detection result from 2D BEV rather than the height information extracted from point cloud data. On the other hand, for the class Pedestrian, the overall BEV IoU value does not show a superior advantage over other views. This indicates that the extraction of dimensional information from the front view and side view for pedestrians is relatively consistent and potentially less challenging compared to the BEV representation. The low IoU score for pedestrians is primarily attributed to their smaller volume, resulting in a significant IoU change with a slight variation of detected coordinates. Even though the utilization of the PCC method has improved the detection of pedestrians in the BEV, detecting pedestrians remains more challenging compared to vehicles due to their smaller BEV area and greater visual complexity. It is also obvious the predicted 3D bounding box that is close to the LiDAR exhibits higher accuracy than that beyond 15 meters.

**Table 2**. Absolute Difference in Coordinates and 3 Dimensions for the Detected 3D Bounding Boxes Against the Ground Truth

| Class | Range (meter) | Absolute Difference in Center Point | | | | | | Absolute Difference of Dimension | | | | | |
|---|---|---|---|---|---|---|---|---|---|---|---|---|---|
| | | X | | Y | | Z | | Length in X | | Length in Y | | Length in Z | |
| | | Value | % | Value | % | Value | % | Value | % | Value | % | Value | % |
| V | Overall | 0.25 | 8.70 | 0.36 | 7.94 | 0.24 | 14.61 | 0.47 | 15.90 | 0.44 | 9.71 | 0.16 | 9.74 |
| | <= 15 | 0.24 | 5.29 | 0.38 | 8.38 | 0.17 | 10.35 | 0.45 | 15.23 | 0.32 | 7.06 | 0.13 | 7.91 |
| | > 15 | 0.31 | 10.49 | 0.48 | 10.59 | 0.26 | 15.83 | 0.48 | 16.24 | 0.69 | 15.22 | 0.26 | 15.83 |
| P | Overall | 0.29 | 17.33 | 0.25 | 14.11 | 0.20 | 11.49 | 0.46 | 27.50 | 0.42 | 23.71 | 0.13 | 7.47 |
| | <= 15 | 0.26 | 15.54 | 0.25 | 14.11 | 0.26 | 14.94 | 0.38 | 22.71 | 0.28 | 15.80 | 0.12 | 6.90 |
| | > 15 | 0.33 | 19.73 | 0.25 | 14.68 | 0.16 | 9.20 | 0.64 | 38.25 | 0.90 | 50.80 | 0.13 | 7.47 |

Notes: V represents Vehicle; P represents Pedestrian

**Table 2** represents the values and percentage of absolute difference of center coordinates and the dimensions along with the X, Y, and Z axis between the predicted bounding boxes and the ground truth, using the average value of ground truth bounding boxes. For both vehicles and pedestrians, the difference in center coordinates of the Z-axis shows a relatively low level of accuracy, suggesting the necessity for a more precise calibration process regarding ground height. Compared with vehicles, the greater percentage of the absolute difference in the center point and three dimensions for pedestrians emphasizes the complexity of the detection process from BEV representation. Moreover, the absolute difference in dimension along the X-axis and Y-axis can indeed be influenced by the low-resolution of the BEV image. The parameters width and length (length in X-axis and Y-axis) of the bounding box are derived from the BEV image, where each pixel is represented using an integer value. This representation can introduce bias when converting the 2D bounding box from the BEV image back into the 3D point cloud data, potentially affecting the accuracy of the dimensions in the X and Y axes.

From **Table 2**, it is evident that the differences in center coordinates and 3 dimensions vary significantly within the specific distance range and beyond, except for the difference in center coordinates along the Z-axis for pedestrians. Despite the less accurate detection of the center point along the Z-axis, the relative height information, represented by the estimated bounding box height, performs comparatively better than the other two dimensions for both vehicles. The variation in the center point along the X-axis and Y-axis is influenced by the object's distance from the LiDAR. When an object is far from the LiDAR, only a portion of its edges can be captured, leading to a smaller area covered by the bounding box in the BEV representation. As the object moves closer to the LiDAR, more complete point cloud data can be collected, resulting in a more accurate bounding box generated from the BEV representation, leading to a precise center point along the X-axis and Y-axis. This result implies that in the context of low-resolution point cloud data, the accuracy of 3D object detection tends to improve as the objects approach closer to the LiDAR device. Additionally, the results suggest that the detection process for pedestrians demands a more complex approach compared to vehicles.

**Evaluation of High-Resolution Parameters**

The high-resolution parameters evaluated in this section include counts, classification, speed, and acceleration. **Table 3** represents the descriptive statistics of speed and acceleration.

**Table 3.** Descriptive Statistics of Speed and Acceleration for Two Classes (Vehicle and Pedestrian)

|  | Vehicle | | | Pedestrian | | |
|---|---|---|---|---|---|---|
|  | Speed (m/s) | Speed (mph) | Acceleration m/s^2 | Speed (m/s) | Speed (mph) | Acceleration m/s^2 |
| Mean | 3.88 | 8.68 | 0.70 | 1.76 | 3.94 | 0.73 |
| Std | 2.64 | 5.90 | 5.26 | 1.80 | 4.02 | 3.93 |
| Min | 0.00 | 0.01 | -25.87 | 0.00 | 0.01 | -23.23 |
| 25% | 1.60 | 3.58 | -0.73 | 0.77 | 1.71 | -0.64 |
| 50% | 3.85 | 8.61 | 0.14 | 1.34 | 3.01 | 0.03 |
| 75% | 5.71 | 12.78 | 1.57 | 1.91 | 4.42 | 1.44 |
| 90% | 7.22 | 16.15 | 5.39 | 3.92 | 8.78 | 4.21 |

Speed Estimation: When calculating the speed of each vehicle, the Gaussian Filter is utilized to eliminate the random fluctuations caused by the location detected. At this signalized intersection, for all detected vehicles, their mean speed is 8.68 mph, which is less than the post speed limit of 20 mph on E Broadway. This average speed includes the entire process, including deceleration upon arrival, waiting during red, and accelerating upon departure; for all detected pedestrians, the calculated average speed of 1.76 m/s and calculated median speed of 1.34 m/s are reasonable compared to the speed provided by Manual on Uniform Traffic Control Devices for Streets and Highways (MUTCD) of 1.2 m/s for the class Average Adult.

Acceleration: The acceleration of detected objects in each frame can be calculated. In future work, this parameter can be used to analyze and predict near-miss crashes.

Classification: This study simply distinguished the classes using the area of bounding boxes in BEV with an accuracy of 84.7%. The False Positive (FP) rate is 0.6%, meaning 0.6% of Pedestrian is incorrectly predicted as Vehicle. This scenario occurred when a crowded group of pedestrians was mistakenly identified as a vehicle during the object recognition process. Such misclassifications can arise due to the complexities involved in distinguishing between densely packed pedestrians and vehicles from the BEV. The False Negative (FN) rate is 14.7%, indicating 14.7% of Vehicle is incorrectly predicted as Pedestrian. The FN rate is relatively higher, primarily because the LiDAR point cloud becomes sparse with increasing distance, leading to a small area of bounding box from BEV that is mistakenly classified as pedestrians.

Counts: Counts from the proposed approach were compared against manual counts at the same intersection. The SAM predicts a total of 103 vehicles, while the manually counted number is 98 vehicles, resulting in an error rate of 5%. However, for pedestrians, the count is 69, while the manual count shows 49, resulting in an error rate of 40.8%. This discrepancy is mainly attributed to the flawed detection results and occluded trajectories in pedestrian counting.

## CONCLUSION

This paper implements a vision system that is able to generate 3D bounding boxes and traffic data from LiDAR point clouds by using a zero-shot learning model and several computer vision techniques. The system first transforms LiDAR point clouds into Bird-Eye View (BEV), followed by the application of a Point Cloud Completion algorithm (PCC) and Segment Anything Model (SAM) to estimate the BEV background and detect moving objects in the foreground respectively. The detected objects from the BEV together with the height information from the perspective view were used to reconstruct a 3D bounding box for each detected object directly after correcting for changing road slopes. Detected objects are subsequently tracked using BYTEtrack to enable traffic parameters to be estimated.

The IOU metric was used to evaluate the accuracy of the proposed framework against ground truth 3D bounding boxes annotated using the Ground Truth Labeler tool in MATLAB. The IOU varied by object class and distance. For vehicles within 15 meters from the LiDAR location, the average 3D volume IOU was about 0.5, and 0.35 for vehicles beyond 15 meters. The IOU score for pedestrians was about 0.25 which is significantly lower compared to vehicles. The low IOU for pedestrians is due to inadequate point cloud density from the single LiDAR used for data collection. The framework was however able to extract high-resolution traffic parameters such as counts, speeds, and acceleration accurately from the LiDAR point cloud data.

Though a robust calibration approach is applied in this framework, there were discrepancies in ground height observed as the distance from the LiDAR device increased. This affected estimation of the center coordinates in Z-axis. To address this issue, future studies should focus on developing precise calibration methods to obtain more reliable ground height information. Additionally, restricting ROI to a specific area can help reduce bias and improve the accuracy of object detection. Moreover, special attention should be directed towards the scaler employed for converting LiDAR point cloud data into a BEV image, as it significantly influences the resolution of the resulting BEV image. It is imperative to recognize that each pixel value in the BEV image must be an integer, which introduces the potential for bias when converting the 2D bounding box from BEV back into the 3D point cloud data. In future work, careful consideration and discussion of appropriate scales are necessary to ensure that each pixel accurately represents a smaller distance, thus improving the precision of both 2D and 3D object detection processes.

## AUTHOR CONTRIBUTIONS

The authors confirm their contribution to the paper as follows: study conception and design: Adu-Gyamfi, Zhang; data collection: Zhang, Yu; data processing: Zhang; analysis and interpretation of results: Zhang, Yu, Aboah, Adu-Gyamfi; draft manuscript: Zhang, Yu, Aboah, Adu-Gyamfi. All Authors reviewed the results and approved the final version of the manuscript.